\documentclass[runningheads]{llncs}

\usepackage[T1]{fontenc}
\usepackage{amsmath}
\usepackage{amssymb}
\usepackage{mathtools}
\usepackage{graphicx}
\usepackage{booktabs}
\usepackage{multirow}
\usepackage{colortbl}
\usepackage{algorithm}
\usepackage{algpseudocode}
\usepackage{bbding}
\usepackage{xcolor}
\usepackage{microtype}
\usepackage{float}
\usepackage[hidelinks]{hyperref}
\usepackage[capitalize,noabbrev]{cleveref}

\makeatletter
\providecommand{\theHALG@line}{}
\renewcommand{\theHALG@line}{\thealgorithm.\arabic{ALG@line}}
\makeatother

\newcommand{\mname}{SEFS}
\newcommand{\methodfull}{Style-Encoder-Free Stylization}

\newcommand{\algorithmblockcaption}[2]{%
  \par\medskip\noindent\refstepcounter{algorithm}\label{#1}%
  \hrule\vspace{2pt}%
  \noindent\textbf{Algorithm~\thealgorithm} #2\par\vspace{2pt}\hrule\vspace{2pt}%
}
\newcommand{\algorithmblockend}{\vspace{2pt}\hrule\par\medskip}

\title{Scale-Separated Conditioning for Style-Encoder-Free Diffusion Stylization}
\titlerunning{Scale-Separated Conditioning for Diffusion Stylization}

\author{Jingtao Zhang\inst{1}\thanks{Equal contribution. Corresponding author.} \and
Haorui Gao\inst{2}\thanks{Equal contribution.} \and
Youqing Liang\inst{2} \and
Zeming Liu\inst{3}}
\authorrunning{J. Zhang et al.}

\institute{
College of Computing, Georgia Institute of Technology, Atlanta, GA, USA\\
\email{jzhang957@gatech.edu}
\and
Courant Institute of Mathematical Sciences, New York University, New York, NY, USA
\and
Department of Computer Science, Brown University, Providence, RI, USA}

\begin{document}
\raggedbottom
\maketitle

\begin{abstract}
Reference-based diffusion stylization requires separating target geometry from transferable appearance. Existing tuning-based methods often rely on aligned content-style-target triplets or auxiliary visual encoders, which increases data cost and can transfer unintended scene structure from the style reference. We propose {\mname} (\methodfull), a style-encoder-free conditioning framework for diffusion transformers. {\mname} forms style tokens from stochastic low-resolution crops of single training images. This crop bottleneck preserves local appearance statistics such as palette, stroke, texture, and material, while reducing access to global layout cues. Target content is encoded by edge and segmentation cues and fused with the noisy latent through parameter-efficient trainable projections. We add style-to-denoising re-normalization for token-statistic alignment and cross-block skip fusion for spatial detail. {\mname} trains on unpaired single images; the frozen diffusion VAE is used only to place image conditions in the latent space. On artistic stylization benchmarks, {\mname} improves content consistency and leakage diagnostics while retaining reference-style affinity, and ablations support the crop-resolution, re-normalization, and skip-fusion choices. The code of {\mname} will be made publicly available.

\end{abstract}

\section{Introduction}
\label{sec:intro}

Diffusion-based text-to-image models have become a common backbone for visual generation, with rapid progress in latent diffusion and diffusion transformers~\cite{ldm,sdxl,sd3,pixartalpha,pixartsigma}. Their pretrained priors support many conditioning tasks, including text prompts, spatial controls, identity references, inpainting or outpainting, and image edits~\cite{controlnet,ipadapter,dreambooth,inpainting,pqdiff,instructpix2pix,imagic,dragdiffusion}. Reference-based stylization is one such conditioning problem. Given a target content image and a style reference, the model must preserve the target geometry while transferring visual appearance such as palette, material, atmosphere, stroke pattern, illumination, and local texture. These attributes are not explicitly labeled in the reference: style is entangled with the reference image's own objects, layout, and scene semantics. This makes stylized data collection, style representation, and evaluation difficult.

This entanglement leads to a common failure mode. If the reference representation is too rich, the model may copy non-transferable structure from the style image, causing content leakage. If it is too weak, the generated image preserves geometry but misses the reference appearance. Existing approaches occupy different points on this trade-off. Training-free methods manipulate inversion trajectories or attention features from the reference~\cite{ddiminversion,styleid,stylealign,deadiff,styleprompt}, which avoids additional training but can be slow when inversion is required, sensitive to inversion quality, and dependent on layer-wise intervention. Adapter-based methods introduce pretrained image encoders or prompt adapters~\cite{ipadapter,styleshot}, but the extracted embeddings are not guaranteed to isolate style from semantics and introduce additional parameters. Tuning-based methods can improve stylization, yet often depend on curated content-style-target triplets or large-scale paired construction~\cite{csgo}, making the data pipeline expensive and limiting scalability.

{\mname} uses a conditioning bottleneck rather than a stronger style encoder. A style reference contains transferable appearance at local scales, whereas its global spatial arrangement is usually non-transferable. We therefore construct style tokens from stochastic low-resolution crops of individual images. Cropping exposes local appearance statistics; resizing suppresses precise contours and global layout; and training across many single images discourages an image-identity shortcut. At inference time, the same style-token pathway is driven by an external style reference, while content is supplied by structural controls from the target image. This creates a same-image-to-cross-image generalization setting: training learns the scale-separated roles from unpaired single images, whereas evaluation tests fixed target-content and external-style-reference pairs.

We propose {\mname} (\methodfull), a style-encoder-free conditioning framework for diffusion transformers. {\mname} separates the conditioning pathway by scale and role: edge maps and segmentation maps provide target-content cues, while low-resolution style crops provide appearance tokens. These conditions are fused into the diffusion transformer with parameter-efficient trainable projections and LoRA adaptation. Style-to-denoising re-normalization aligns the statistics of style tokens with the denoising token stream, and cross-block skip fusion reuses shallow structural features in later transformer blocks. The framework uses unpaired single images for training, avoids auxiliary style encoders, and remains parameter-efficient.

Our contributions are summarized as follows:

\textbf{i) }We introduce {\mname}, a style-encoder-free diffusion stylization framework that learns from unpaired single images rather than aligned content-style-target triplets. The method uses a scale-separated crop bottleneck to preserve local appearance statistics while suppressing global reference structure.

\textbf{ii) }We design a parameter-efficient trainable conditioning architecture that combines target edges, segmentation maps, and low-resolution style tokens inside a diffusion transformer. Cross-block skip fusion improves content preservation by carrying shallow structural information to later denoising stages.

\textbf{iii) }We introduce style-to-denoising re-normalization to stabilize heterogeneous token fusion. This module improves the crop-token pathway in our ablations and supports training using about 40k unpaired WikiArt images, without requiring curated triplets or a separate style-encoder training set.

\textbf{iv) }We evaluate {\mname} against recent diffusion stylization methods and conduct focused ablations on content conditions, crop resolution, style-encoder usage, re-normalization, and skip fusion. The results improve content scores and leakage diagnostics while retaining reference-style affinity.

\section{Related Work}
\begin{figure}[tb!]
    \centering
    \includegraphics[width=1\linewidth]{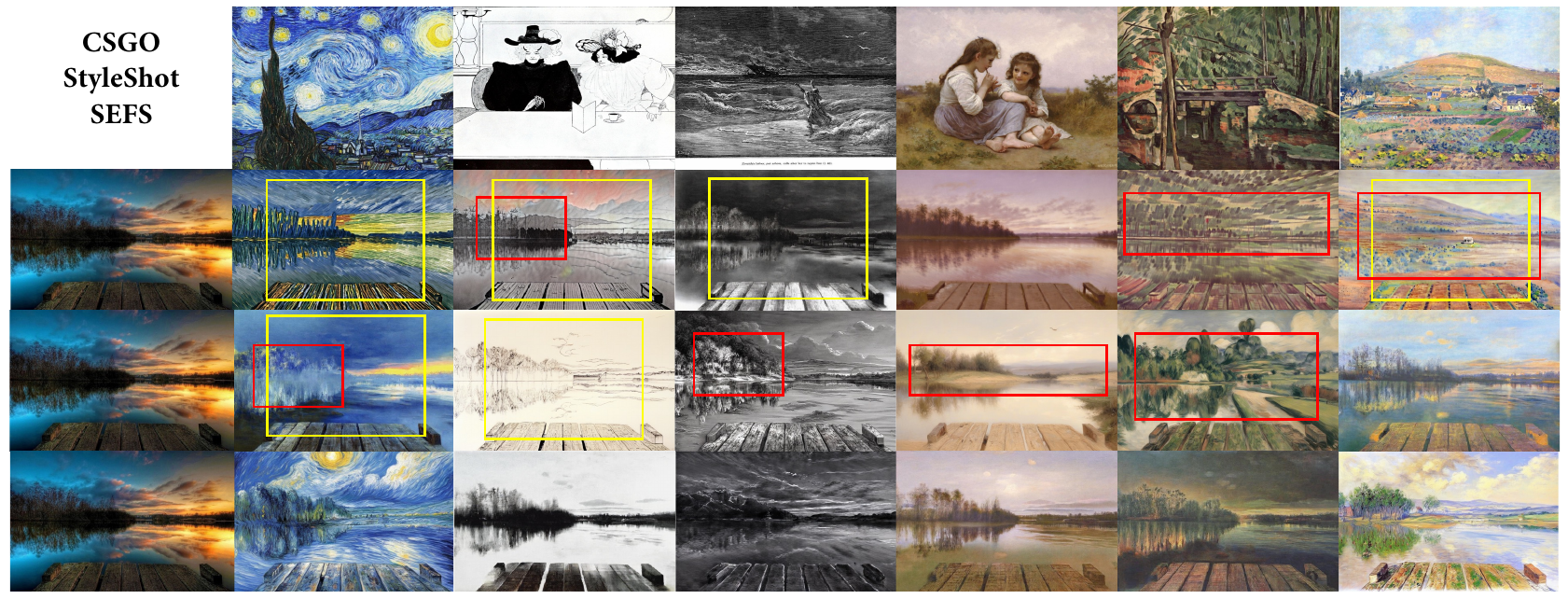}
    \vspace{-15pt}
    \caption{Representative comparisons with CSGO~\cite{csgo}, StyleShot~\cite{styleshot}, and {\mname}. Existing methods can drift from the target content or transfer reference-specific structure; scale-separated conditioning helps {\mname} preserve target layout while matching reference appearance.}
    \label{fig:fig2}
\end{figure}

\subsection{Diffusion Models and Conditional Generation}
Latent diffusion models~\cite{ldm,sdxl} and diffusion transformers~\cite{dit,sd3,pixartalpha,pixartsigma,roleawarevideo,swiftvideo} provide strong priors for controllable image synthesis. Typical conditioning methods keep the generator largely fixed and add spatial adapters~\cite{controlnet}, image-prompt adapters~\cite{ipadapter}, or low-rank tuning~\cite{lora}. This makes adaptation efficient but creates a representation trade-off: conditions must guide generation without overriding target content or leaking unwanted semantics. In stylization, spatial maps preserve geometry but not appearance, whereas image embeddings can entangle style with reference identity, objects, and layout. {\mname} keeps the parameter-efficient paradigm but replaces the auxiliary style encoder with a scale-separated image pathway.

\subsection{Reference-Based Stylization}
Arbitrary reference-based stylization remains difficult because style may include color, material, atmosphere, stroke statistics, and sometimes global structure. Classical neural style transfer separates content and style through feature statistics, as in AdaIN-style normalization~\cite{adain}, while transformer feed-forward methods improve perceptual quality and content preservation~\cite{stytr2}. Diffusion methods exploit stronger generative priors, but still need a style pathway that transfers appearance without copying reference layout.

Recent diffusion stylization methods are commonly training-free, adapter-based, or tuning-based. Training-free approaches manipulate inversion trajectories, attention features, reference modulation, or sampling startpoints~\cite{ddiminversion,styleid,stylealign,deadiff,styleprompt}. They avoid task-specific training but can be sensitive to inversion quality, layer choices, and attention reuse. Adapter-based methods use image-conditioned pathways or pretrained visual encoders~\cite{ipadapter,styleshot}, which capture rich appearance but may retain object category, layout, or reference identity. Tuning-based methods such as CSGO~\cite{csgo} improve stylization end-to-end, often with curated triplets or large construction pipelines.

{\mname} is complementary to these directions. Instead of extracting a full-resolution style embedding with a pretrained encoder, it forms style tokens from stochastic low-resolution crops. This bottleneck preserves local transferable appearance while suppressing global reference geometry; combined with target structural cues and token re-normalization, it yields a diffusion-transformer stylization framework trained without triplet supervision or an auxiliary visual style encoder.

\section{The proposed {\mname}}
\label{sec:method}

\begin{figure}[tb!]
    \centering
    \includegraphics[width=1.\linewidth]{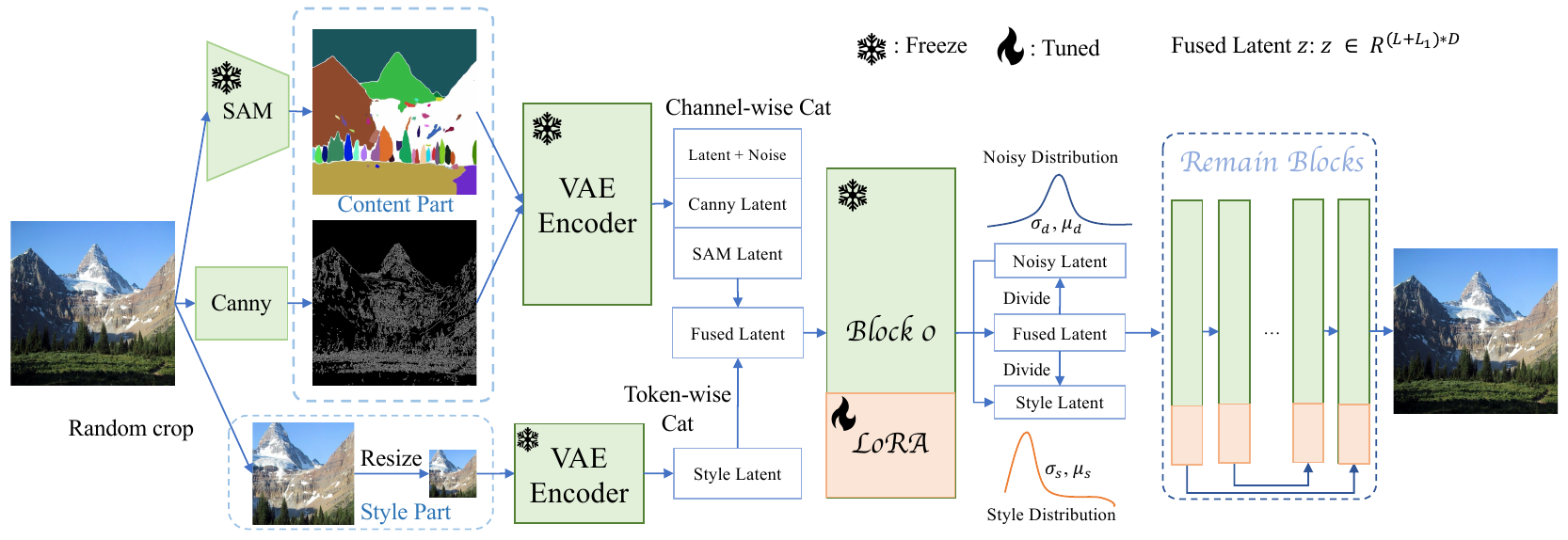}
    \vspace{-20pt}
    \caption{Framework of the proposed {\mname}. The target image provides structural content conditions through Canny edges and SAM~\cite{sam} segmentation, while the style pathway receives stochastic low-resolution crops that expose local appearance statistics and suppress global layout. Style tokens are concatenated with the denoising tokens and re-normalized using the denoising-token statistics $(\mu_{\mathbf{d}},v_{\mathbf{d}})$ and style-token statistics $(\mu_{\mathbf{s}},v_{\mathbf{s}})$. The right part illustrates LoRA fine-tuning and cross-block skip fusion.}
    \label{fig:framework}
\end{figure}

\subsection{Preliminaries}
\textbf{Flow Matching.} Let $p_0$ and $p_1$ denote the source and data distributions on $\mathbb{R}^{n}$. An ODE-based generative model transports a sample $\mathbf{x}_0\sim p_0$ to $\mathbf{x}_1\sim p_1$ by integrating a time-dependent vector field:

\begin{equation}
\label{eqn:ode}
\frac{d\mathbf{x}_t}{dt}=f_\theta(\mathbf{x}_t,t),\quad t\in[0,1].
\end{equation}
The learned vector field $f_\theta:\mathbb{R}^{n}\times[0,1]\rightarrow\mathbb{R}^{n}$ defines the trajectory followed during generation.

\textbf{Rectified Flow. }Rectified Flow~\cite{reflow} trains the vector field on straight paths between a data sample and Gaussian noise. We use an equivalent reverse-time parameterization of the SD3 latent-space path~\cite{sd3}, where $t=0$ corresponds to noise and $t=1$ corresponds to the data latent:
\begin{equation}
\label{eqn:rectified_flow}
\mathbf{z}_{t} = (1-t)\epsilon + t\mathbf{z}_0,\quad \epsilon\sim\mathcal{N}(0,I),
\end{equation}
and minimize
\begin{equation}
\label{eqn:rf_loss}
\mathcal{L}_{\mathrm{RF}} = \mathbb{E}_{t,\epsilon}\left[\lambda_t\left\|f_\theta(\mathbf{z}_t,t)-(\mathbf{z}_0-\epsilon)\right\|_2^2\right],
\end{equation}
where $\lambda_t$ is the timestep-dependent loss weight.

\begin{figure}[tb!]
    \centering
    \includegraphics[width=.9\linewidth]{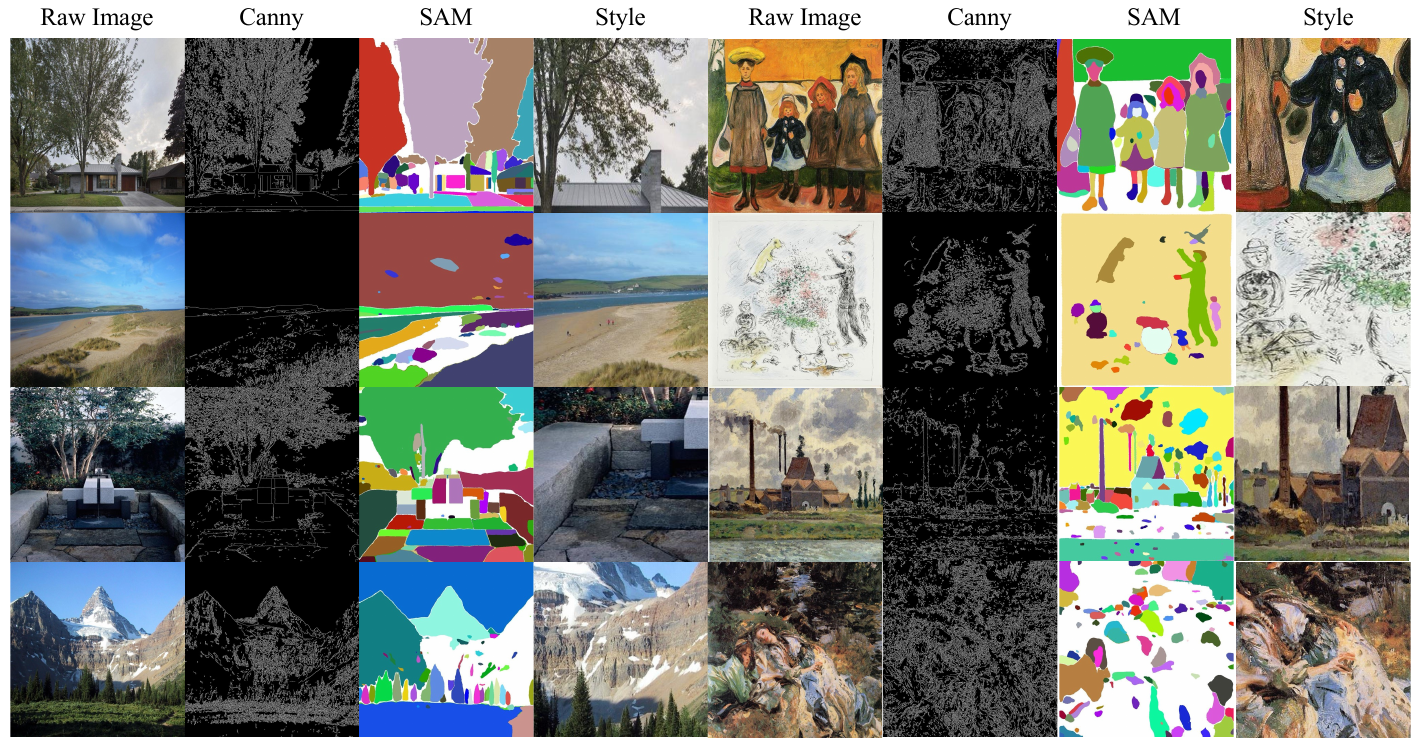}
    \vspace{-10pt}
    \caption{Training inputs of {\mname}. Canny edges preserve fine structural boundaries, SAM masks preserve coarse region layout, and low-resolution style crops retain local appearance while reducing access to global reference geometry.}
    \label{fig:training_sample}
\end{figure}

\subsection{The proposed {\mname}}
We first describe how {\mname} constructs content and style conditions from single images, then present the token projection, re-normalization, cross-block skip fusion, and training objective.

\textbf{Condition Generation. }Given a training image $\mathbf{x}$, {\mname} constructs a pseudo content-style training instance without requiring an external stylized target. For content, we use two complementary structural views. Edge maps such as Canny~\cite{canny} and HED~\cite{hed} are commonly used to provide high-frequency boundaries and local contours; we instantiate this view with Canny edges. SAM segmentation masks~\cite{sam} provide region-level organization that is not captured by sparse edges alone, such as consistent regions over buildings, sky, or mountains. For style, we apply stochastic cropping to $\mathbf{x}$ and resize the crop to a low resolution. This operation preserves local appearance statistics such as color palette, texture, material, and stroke pattern, but removes much of the precise contour and global layout information that would otherwise encourage content leakage. During inference, the cropped training view is replaced by the user-provided style reference, processed through the same low-resolution style pathway.

\begin{figure}
    \centering
    \includegraphics[width=\linewidth]{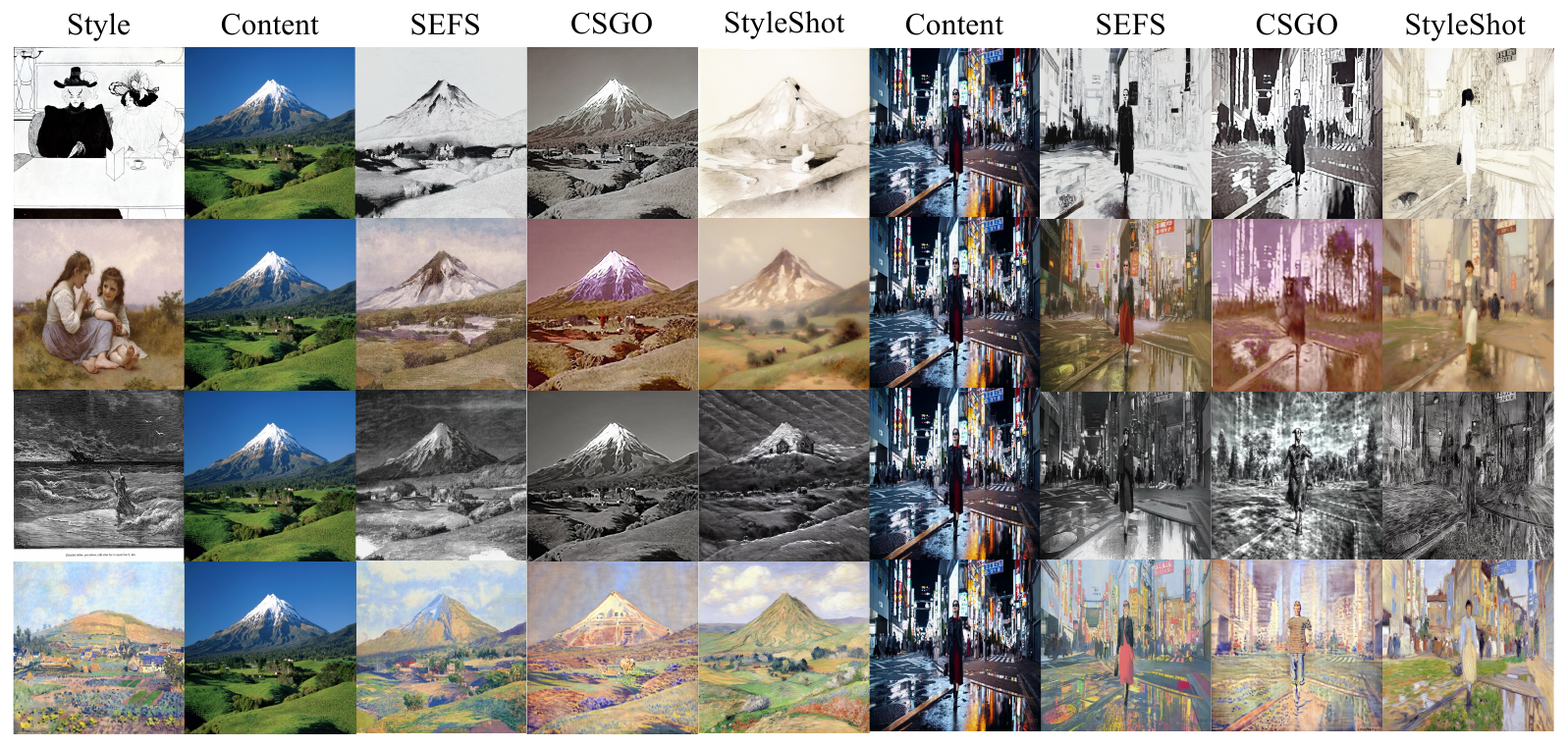}
    \vspace{-10pt}
    \caption{Qualitative comparisons. {\mname} better preserves target layout while transferring reference appearance; baselines either weaken style transfer or introduce reference-induced content drift.}
    \label{fig:main}
\end{figure}

Implementation details. During training, $RandomCropResize$ samples a crop with ratio $(0.1\sim0.5)$ from the training image and resizes it to $64\times64$ to form the style view. During inference, we use a deterministic single-view protocol: the full style reference is resized to the same $64\times64$ style-token resolution rather than sampled repeatedly. Canny maps and SAM outputs are rendered as three-channel images before VAE encoding, matching the input interface of the copied SD3 patch-embedding branch. Text-conditioning settings are kept fixed across training and evaluation so that differences primarily reflect the image-conditioning pathways. Although training constructs pseudo instances from a single image, the content and style signals are deliberately separated by role and scale: structural maps provide geometry, while the low-resolution style view provides appearance statistics with limited access to global layout. The cross-image evaluations in Fig.~\ref{fig:main} and Table~\ref{tab:main_compare} therefore test whether this bottleneck transfers to arbitrary content-style pairs rather than only reconstructing the training image.

\textbf{Condition projection. }Let $\mathbf{m}\in\mathbb{R}^{H\times W \times 3}$ and $\mathbf{c}\in\mathbb{R}^{H\times W \times 3}$ denote the SAM segmentation map and Canny edge map, respectively. Let $\mathbf{s}\in\mathbb{R}^{H_1\times W_1\times 3}$ denote the low-resolution style view, where $H_1 < H$ and $W_1 < W$. We use pretrained SD3~\cite{sd3} as the base model and form the noisy latent by
\begin{equation}
\label{eqn:add_noise}
\mathbf{z}_{t} = (1-t) \cdot \epsilon + t \cdot \mathbf{z}_0
\end{equation}
where $\mathbf{z}_0=Encode(\mathbf{x})$ and $Encode(\cdot)$ is the pretrained VAE encoder. We encode the content and style conditions in the same latent space: $\mathbf{m}_{\mathbf{z}}=Encode(\mathbf{m})$, $\mathbf{c}_{\mathbf{z}}=Encode(\mathbf{c})$, and $\mathbf{s}_{\mathbf{z}}=Encode(\mathbf{s})$. Thus, ``style-encoder-free'' means that {\mname} does not introduce an auxiliary CLIP/IP-Adapter-style visual encoder for style extraction; all image conditions are encoded only through the frozen diffusion VAE before entering the parameter-efficient trainable conditioning branch. After patchification with patch size $p$ and projection by the copied SD3 patch-embedding layer, the denoising latent, segmentation latent, and edge latent are represented as sequences of length $L$ and hidden dimension $D$, while the style latent has length $L_1$. We concatenate content-related tokens along the channel dimension:
\begin{equation}
    \mathbf{z}_{cat} = Concat (\mathbf{z}_{t},\mathbf{m}_{\mathbf{z}},\mathbf{c}_{\mathbf{z}})
\end{equation}
where $\mathbf{z}_{cat}\in\mathbb{R}^{L\times 3D}$. A trainable projection $\mathbf{W}_{cat}\in\mathbb{R}^{3D\times D}$ maps the concatenated sequence back to the original hidden dimension. We initialize this projection to preserve the pretrained denoising stream at the beginning of fine-tuning:
\begin{equation}
\label{eqn:patchify_linear}
\mathbf{W}_{cat} = \\
\begin{pmatrix}
1 & 0 & \cdots & 0 \\
0 & 1 & \cdots & 0 \\
\cdots & \cdots & \cdots & \cdots \\
0 & 0 & \cdots & 1 \\
0 & 0 & 0 & 0 \\
\cdots & \cdots & \cdots & \cdots \\
0 & 0 & 0 & 0 \\
\end{pmatrix}, \ \ \mathbf{z} = \mathbf{z}_{cat} \mathbf{W}_{cat}
\end{equation}
where $\mathbf{W}_{cat}$ is initialized by concatenating one $D\times D$ identity matrix and two $D\times D$ zero matrices. The style tokens are then concatenated with the projected denoising sequence along the token dimension, yielding $\mathbf{z}\in\mathbb{R}^{(L+L_1)\times D}$.

\textbf{Re-normalization. }The denoising tokens and style tokens originate from different visual views and therefore have different activation statistics. Direct token concatenation can make the style stream either dominate generation or be ignored during early fine-tuning. To stabilize the fusion, after the first transformer block we divide the sequence into the denoising part $\mathbf{d}_{z}=\mathbf{z}[0:L]$ and the style part $\mathbf{s}_{z}=\mathbf{z}[L:L+L_1]$. We then align the style-token statistics to the denoising-token statistics:
\begin{equation}
\label{eqn:renorm}
\tilde{\mathbf{s}}_{\mathbf{z}} =
\frac{\sqrt{v_{\mathbf{d}}+\eta}}{\sqrt{v_{\mathbf{s}}+\eta}}
(\mathbf{s}_{\mathbf{z}}-\mu_{\mathbf{s}})+\mu_{\mathbf{d}}
\end{equation}
where $\mu_{\mathbf{s}}$ and $v_{\mathbf{s}}$ are the channel-wise mean and variance of the style tokens, $\mu_{\mathbf{d}}$ and $v_{\mathbf{d}}$ are the corresponding statistics of the denoising tokens, and $\eta$ is a small constant for numerical stability. The square roots convert variances into standard-deviation scales for feature-statistic matching. Similar to adaptive modulation in diffusion transformers~\cite{dit}, we predict a residual scale and shift from the re-normalized style tokens:
\begin{equation}
    \boldsymbol{\gamma}_{\mathbf{s}}=\tilde{\mathbf{s}}_{\mathbf{z}}\mathbf{W}_{scale},\quad
    \boldsymbol{\beta}_{\mathbf{s}}=\tilde{\mathbf{s}}_{\mathbf{z}}\mathbf{W}_{shift},\quad
    \mathbf{s}_{\mathbf{z}} =
    \tilde{\mathbf{s}}_{\mathbf{z}} \odot (1+\boldsymbol{\gamma}_{\mathbf{s}})
    + \boldsymbol{\beta}_{\mathbf{s}}
\end{equation}
where $\mathbf{W}_{scale}\in\mathbb{R}^{D\times D}$ and $\mathbf{W}_{shift}\in\mathbb{R}^{D\times D}$ are trainable parameters.

\textbf{Skip-Connection across blocks. }Transformer blocks at shallow depths preserve fine structural details, while deeper blocks tend to carry more semantic and global information. To improve content preservation, we introduce cross-block skip fusion in the latter half of the transformer. Let $\mathbf{z}^{(i)}$ denote the output sequence of block $i$ in a model with $N$ blocks. For later blocks, we concatenate the current sequence with its paired shallow feature along the channel dimension and project it back to $D$ dimensions:
\begin{equation}
\label{eqn:skip_init}
\mathbf{W}_{skip}^{(i)} = \\
\begin{pmatrix}
1 & 0 & \cdots & 0 \\
0 & 1 & \cdots & 0 \\
\cdots & \cdots & \cdots & \cdots \\
0 & 0 & \cdots & 1 \\
0 & 0 & 0 & 0 \\
\cdots & \cdots & \cdots & \cdots \\
0 & 0 & 0 & 0 \\
\end{pmatrix}, \ \ \mathbf{z}^{(i)} = \mathbf{z}^{(i)} \mathbf{W}_{skip}^{(i)}
\end{equation}
The skip projection is initialized to pass through the current block output and gradually learn how much shallow structural information to reuse. The resulting sequence $\mathbf{z}^{(i)}\in\mathbb{R}^{(L+L_1)\times D}$ is fed into the next block.

\textbf{Objective function. }The training objective follows the SD3 rectified-flow loss with additional conditions:
\begin{equation}
    \mathcal{L}_{\mname} = \lambda_t \cdot \Vert f(\mathbf{z}_t,\mathbf{m}_{\mathbf{z}},\mathbf{c}_{\mathbf{z}},\mathbf{s}_{\mathbf{z}}, t; \theta) - (\mathbf{z}_0-\epsilon) \Vert_2^2
\end{equation}
where $\lambda_t$ balances different timesteps.

\textbf{Algorithm summary. }We summarize the training pipeline in Algorithm~\ref{alg:training} and the block-wise forward pass in Algorithm~\ref{alg:block}. Algorithm~\ref{alg:training} shows how each unpaired image is converted into structural content conditions and a low-resolution style view, encoded into latent tokens, and optimized with the rectified-flow objective. Algorithm~\ref{alg:block} details the corresponding forward computation, including content-token projection, style-token re-normalization, and cross-block skip fusion. Here, $ConCat_{Dim}$ and $ConCat_{Seq}$ denote concatenation over the channel dimension and token dimension, respectively; at inference, the random crop is replaced by the user-provided style reference while the rest of the conditioning pipeline is unchanged.

\algorithmblockcaption{alg:training}{Training pipeline of the proposed {\mname}. $RandomCropResize$ denotes stochastic cropping followed by low-resolution resizing.}
\begin{algorithmic}[1]
\State \textbf{Input: }Pretrained SD3 model $f_{\theta}$. Raw image $\mathbf{x}$, text embeddings $\mathbf{T}$. Pretrained SAM encoder $SAM(\cdot)$ and VAE encoder $Encode(\cdot)$. Canny detector $Canny(\cdot)$, and Optimizer $OPT(\cdot)$.
\State \textbf{Init: }$\Theta_{\mathrm{train}}$ including LoRA, projection, re-normalization, and skip-fusion parameters
\State \textbf{Data:} $\mathbf{s} \leftarrow RandomCropResize(\mathbf{x})$
\State \textbf{Data:} $\mathbf{m} \leftarrow SAM(\mathbf{x})$, $\mathbf{c} \leftarrow Canny(\mathbf{x})$
\State \textbf{Encode:} $\mathbf{m}_{\mathbf{z}}, \mathbf{c}_{\mathbf{z}}, \mathbf{s}_{\mathbf{z}}, \mathbf{z}_{0} \leftarrow Encode(\mathbf{m}, \mathbf{c}, \mathbf{s}, \mathbf{x})$
\State \textbf{Diffusion:} $\mathbf{z}_{t} = (1-t) \cdot \epsilon + t \cdot \mathbf{z}_{0}$, $\epsilon \sim \mathcal{N}(0, \mathbf{I})$
\State \textbf{Forward:} $\mathcal{L}_{\mname} = \Vert f_{\theta}(\mathbf{z}_{t}, \mathbf{m}_{\mathbf{z}}, \mathbf{c}_{\mathbf{z}}, \mathbf{s}_{\mathbf{z}}) - (\mathbf{z}_0-\epsilon)\Vert_2^2$
\State \textbf{Backward:} $\Theta_{\mathrm{train}} = OPT(\partial \mathcal{L}_{\mname} / \partial \Theta_{\mathrm{train}})$
\State \textbf{Return: }SD3 model with trained LoRA and trainable conditioning modules
\end{algorithmic}
\algorithmblockend

\newpage
\algorithmblockcaption{alg:block}{Block-wise forward of the proposed {\mname}.}
\begin{algorithmic}[1]
\State \textbf{Input: }$i$-th block with LoRA weights $Block^{(i)}(\cdot), 0\leq i \leq N-1$, and $N$ is the number of blocks. Noisy image latent $\mathbf{z}_{t}$, style image latent $\mathbf{s}_{\mathbf{z}}$, Canny image latent $\mathbf{c}_{\mathbf{z}}$, and SAM image latent $\mathbf{m}_{\mathbf{z}}$. Extra learnable parameters $\mathbf{W}_{cat}$ and $\mathbf{W}_{skip}^{(i)}, \frac{N}{2}\leq i \leq N-1$. Shift and scale $\mathbf{W}_{shift}$, $\mathbf{W}_{scale}$.
\State \textbf{Init:} $j \leftarrow 0$, $skipList=[]$
\State \textbf{Patchify: }$\mathbf{z}_{t},\mathbf{s}_{\mathbf{z}}, \mathbf{c}_{z}, \mathbf{m}_{\mathbf{z}}  =Patchify(\mathbf{z}_{t},\mathbf{s}_{\mathbf{z}}, \mathbf{c}_{z}, \mathbf{m}_{\mathbf{z}})$
\State \textbf{Content Fusion: }$\mathbf{z} = ConCat_{Dim}(\mathbf{z}_{t}, \mathbf{c}_{z}, \mathbf{m}_{\mathbf{z}}) \mathbf{W}_{cat}$
\State \textbf{Style Fusion: }$\mathbf{z}^{(0)} = ConCat_{Seq}(\mathbf{z}, \mathbf{s}_{\mathbf{z}})$

\For{$j\leq N-1$}
    \If{$j == 0$}
        \State \textbf{Store: }$skipList.append(\mathbf{z}^{(j)})$
        \State \textbf{Forward: }$\mathbf{z}^{(j+1)}= Block^{(j)}(\mathbf{z}^{(j)})$
        \State \textbf{Divide: }$\mathbf{d}_{\mathbf{z}}, \mathbf{s}_{\mathbf{z}} \leftarrow Divide_{Seq}(\mathbf{z}^{(j+1)})$
        \State \textbf{Statistics: }$\mu_{\mathbf{d}},v_{\mathbf{d}}\leftarrow Stats(\mathbf{d}_{\mathbf{z}})$
        \State \textbf{Statistics: }$\mu_{\mathbf{s}},v_{\mathbf{s}}\leftarrow Stats(\mathbf{s}_{\mathbf{z}})$
        \State \textbf{Standardization: }$\mathbf{s}_{z} \leftarrow \frac{\mathbf{s}_{\mathbf{z}} - \mu_{\mathbf{s}}}{\sqrt{v_\mathbf{s} + \eta}}, \eta=10^{-5}$
        \State \textbf{Re-Normalization: }$\tilde{\mathbf{s}}_{\mathbf{z}} \leftarrow \mathbf{s}_{z} \odot \sqrt{v_{\mathbf{d}}+\eta} + \mu_{\mathbf{d}}$
        \State \textbf{Scale/Shift: }$\mathbf{a}_{\mathbf{s}}\leftarrow 1+\tilde{\mathbf{s}}_{\mathbf{z}}\mathbf{W}_{scale}$
        \State \textbf{Scale/Shift: }$\mathbf{b}_{\mathbf{s}}\leftarrow \tilde{\mathbf{s}}_{\mathbf{z}}\mathbf{W}_{shift}$
        \State \textbf{Re-Scale: }$\mathbf{s}_{\mathbf{z}} \leftarrow \tilde{\mathbf{s}}_{\mathbf{z}}\odot\mathbf{a}_{\mathbf{s}}+\mathbf{b}_{\mathbf{s}}$
        \State \textbf{Style-Fusion: }$\mathbf{z}^{(j+1)} \leftarrow Cat_{Seq}(\mathbf{d}_{\mathbf{z}}, \mathbf{s}_{\mathbf{z}})$
    \ElsIf{$j\leq (N-1) / 2$}
        \State \textbf{Store: }$skipList.append(\mathbf{z}^{(j)})$
        \State \textbf{Forward: }$\mathbf{z}^{(j+1)}= Block^{(j)}(\mathbf{z}^{(j)})$
    \Else
        \State \textbf{Skip:} $\mathbf{z}^{(j)} = Cat_{Dim}(\mathbf{z}^{(j)}, skipList[N-1-j])$
        \State \textbf{Fusion:} $\mathbf{z}^{(j)} \leftarrow \mathbf{z}^{(j)}\mathbf{W}^{(j)}_{skip}$
        \State \textbf{Forward: }$\mathbf{z}^{(j+1)}= Block^{(j)}(\mathbf{z}^{(j)})$
    \EndIf
    \State $j = j+1$
    \EndFor
    \State \Return $\mathbf{z}_{(N)}$
\end{algorithmic}
\algorithmblockend

\textbf{Inference pipeline. }Given a target content image $\mathbf{r}$ and a style reference image $\mathbf{s}$, we compute the Canny edge map $\mathbf{c}$ and SAM segmentation map $\mathbf{m}$ from $\mathbf{r}$. The style reference is passed through the same low-resolution style pathway used during training. We then encode $\mathbf{m}$, $\mathbf{c}$, and $\mathbf{s}$ into latent conditions $\mathbf{z}_{\mathbf{m}}$, $\mathbf{z}_{\mathbf{c}}$, and $\mathbf{z}_{\mathbf{s}}$, and denoise a randomly initialized latent conditioned on all three signals.

\section{Experimental Results}
\textbf{Experimental setup. }We use SD3-Medium as the base generator and train only LoRA layers, content/style projections, re-normalization scale/shift, and cross-block skip projections. A copied SD3 patch-embedding layer encodes $\mathbf{s}_{z}$, $\mathbf{c}_{z}$, and $\mathbf{m}_{z}$; the original branch remains frozen.

Training uses WikiArt~\cite{wikiart}, AdamW~\cite{adamw}, 30k steps, learning rate $2\times10^{-5}$, weight decay $0.03$, batch size 32, and 8 A100 GPUs. \textbf{Evaluation protocol.} The main comparison uses 1,000 fixed cross-image content-style pairs against StyleShot~\cite{styleshot} and CSGO~\cite{csgo}. Canny/SAM are computed only from the target content image; style references enter through each method's style-conditioning input. All methods use official checkpoints or recommended preprocessing, generate $512\times512$ images with 30 steps, and share prompts, seeds, and sampler settings. We use LoRA rank 16, SAM ViT-H masks, and Canny thresholds 100/200. Content DINO and CLIP-I measure target consistency; Style Ref. Sim. is CLIP image-image similarity to the reference; Leakage is a DINO diagnostic between edge-rendered outputs and style references, where lower is better; FID is computed against held-out WikiArt style-domain images. Human preference is the primary aggregate signal, and automatic metrics are diagnostics. Pseudo-triplet ablations in Tables~\ref{tab:content} and~\ref{tab:style} are used only for mechanism analysis.

\begin{table}[H]
    \centering
    \caption{Content-condition and skip-fusion ablation on WikiArt. All variants use 30k steps, batch size 32, and learning rate $2\times10^{-5}$.}
    \vspace{-5pt}
    \resizebox{\linewidth}{!}{\begin{tabular}{c|c|c|c|c|c}
        \toprule[1pt]
        Skip-connection & SAM & Canny & IS & FID & CLIP-I \\
        \midrule[0.8pt]
        \Checkmark & \Checkmark & \Checkmark & \textbf{9.1788} & \textbf{48.7677} & \textbf{0.9063} \\
        \XSolidBrush & \Checkmark & \Checkmark & 9.1304 & 53.1521 & 0.8813 \\
        \Checkmark & \XSolidBrush & \Checkmark & 9.1543 & 49.3103 & 0.8983 \\
        \Checkmark & \Checkmark & \XSolidBrush & 9.0941 & 59.1620 & 0.8294 \\
        \bottomrule[1pt]
    \end{tabular}}
    \label{tab:content}
\end{table}

\subsection{Main Comparisons}
Fig.~\ref{fig:main} compares {\mname} with CSGO~\cite{csgo} and StyleShot~\cite{styleshot}. Across diverse pairs, {\mname} preserves target structure while transferring palette, texture, and stroke statistics. The references contain non-transferable foreground figures, village layouts, and line-art contours; baselines more often inherit these structures or distort target geometry.

Table~\ref{tab:main_compare} reports the main comparison. Standard baselines use their released pipelines. To isolate the content-control confound, StyleShot + Struct. Ctrl. keeps the StyleShot style encoder and injection pathway, adds the same target-derived Canny/SAM branch as {\mname}, and trains only the added structural projection and LoRA layers with the same data, rank, and step budget. It does not use the SEFS low-resolution style-token pathway or re-normalization.

\begin{table}[H]
    \centering
    \caption{Main comparison. StyleShot + Struct. Ctrl. gives StyleShot the same target-derived Canny/SAM controls as {\mname}. Time is per image on A100. Lower Leakage/FID and higher content/style scores are better.}
    \vspace{-5pt}
    
    \resizebox{\linewidth}{!}{\begin{tabular}{c|c|c|c|c|c|c}
        \toprule[1pt]
        Method & Content DINO & Content CLIP-I & Style Ref. Sim. & Leakage & FID & Time \\
        \midrule[0.8pt]
        StyleShot & 0.8017 & 0.8684 & 0.7462 & 0.4871 & 56.8421 & 4.8s \\
        StyleShot + Struct. Ctrl. & 0.8918 & 0.8971 & 0.7785 & 0.4472 & 51.9248 & 7.2s \\
        CSGO & 0.8098 & 0.8566 & 0.7397 & 0.4944 & 60.6377 & 6.3s \\
        {\mname} & 0.9040 & 0.9026 & 0.7835 & 0.3931 & 49.5724 & 6.1s \\
        \bottomrule[1pt]
    \end{tabular}}
    \label{tab:main_compare}
\end{table}

The matched-control baseline narrows the content gap, confirming the value of structural conditions, but still has higher Leakage and worse FID despite similar content and Style Ref. Sim. scores. Thus, {\mname}'s gain is not explained by Canny/SAM alone; under matched structure control, the low-resolution style bottleneck better suppresses non-transferable reference layout.

\textbf{Leakage validation. }We annotate 100 hard pairs with non-transferable objects or layouts. The DINO edge-rendering score correlates with human leakage ratings with Spearman $\rho=0.62$, while Style Ref. Sim. is weaker ($\rho=0.27$), supporting Leakage as an auxiliary diagnostic. \textbf{Efficiency. }{\mname} uses no auxiliary learned visual style encoder; image conditions pass through the frozen VAE, and online structural preprocessing is counted in end-to-end latency. It adds 42.7M trainable parameters, takes 3.9s model-only time on A100, and takes 6.1s end-to-end with online Canny/SAM. \textbf{Human preference. }In blind forced-choice comparisons, 42 participants judged six content-style pairs per comparison, yielding 252 judgments for each method pair. Each question showed the same content image, style reference, and two anonymized outputs in random order. Under the joint criterion of content preservation and style fidelity, participants preferred {\mname} over StyleShot in 64.3\% of comparisons (95\% Wilson CI: 58.2--69.9\%), over CSGO in 72.2\% (95\% Wilson CI: 66.4--77.4\%), and over StyleShot + Struct. Ctrl. in 62.7\% (95\% Wilson CI: 56.6--68.4\%).

Table~\ref{tab:mechanism_control} gives cross-image mechanism controls. Removing or shuffling style tokens preserves content but reduces style-reference similarity, showing that the style pathway carries transferable appearance. SEFS w/ Style Encoder keeps the same SD3 backbone, content branch, LoRA rank, data, and budget, but replaces low-resolution VAE style tokens with a frozen CLIP visual style encoder and trainable projection.

\begin{table}[H]
    \centering
    \caption{Cross-image mechanism controls on the same 1,000 content-style pairs.}
    \vspace{-5pt}
    
    \resizebox{\linewidth}{!}{\begin{tabular}{c|c|c|c|c}
        \toprule[1pt]
        Variant & Content DINO & Style Ref. Sim. & Leakage & FID \\
        \midrule[0.8pt]
        Full SEFS & 0.9040 & 0.7835 & 0.3931 & 49.5724 \\
        SEFS w/ Style Encoder & 0.8924 & 0.7946 & 0.4478 & 52.3861 \\
        w/o style tokens & 0.9132 & 0.7048 & 0.3619 & 55.2846 \\
        shuffled style & 0.9015 & 0.7162 & 0.3698 & 53.9173 \\
        w/o re-normalization & 0.8618 & 0.7447 & 0.4326 & 58.6035 \\
        high-res style tokens & 0.8654 & 0.7779 & 0.4612 & 55.9184 \\
        \bottomrule[1pt]
    \end{tabular}}
    \label{tab:mechanism_control}
\end{table}

The architecture-matched control reaches slightly higher Style Ref. Sim. than {\mname}, but also increases Leakage and FID, supporting the trade-off between reference affinity and reference-layout leakage.

\subsection{Ablation Studies}
\textbf{Content Consistency. }We sample 1,000 WikiArt test images and construct pseudo triplets with SAM, Canny~\cite{canny}, and a $64\times64$ random style crop, isolating content conditioning because Canny and SAM encode structure rather than appearance. For Canny-only or SAM-only variants, $\mathbf{W}_{cat}\in\mathbb{R}^{2D\times D}$ matches the input dimension.

Table~\ref{tab:content} shows that the full model performs best. Removing Canny causes the largest FID and CLIP-I degradation, highlighting edge-level geometry; removing SAM has a smaller but consistent effect on coarse layout. Skip fusion further improves FID and CLIP-I.

\textbf{Style Consistency. }We analyze re-normalization and crop resolution with 10k-step variants. For the resolution ablation, the style-view resolution is varied during training. At evaluation time, all style views are first downsampled to $64\times64$ and then resized to the corresponding training resolution, so the comparison tests the learned resolution-specific pathway rather than giving higher-resolution variants extra reference information. The encoder variant replaces crop tokens with a pretrained CLIP visual encoder and trainable projection.

\begin{table}[H]
    \centering
    \caption{Style-token ablation on WikiArt. ``Style Res'' is the style-view resolution; ``Style Enc'' replaces crop tokens with a pretrained CLIP visual encoder.}
    \vspace{-5pt}
    \resizebox{\linewidth}{!}{\begin{tabular}{c|c|c|c|c|c}
        \toprule[1pt]
        Re-Normalization & Style Res & Style Enc & IS & FID & CLIP-I \\
        \midrule[0.8pt]
        \Checkmark & $64\times 64$ & \XSolidBrush & \textbf{9.4697} & \textbf{54.1585} & \textbf{0.8821} \\
        \XSolidBrush & $64\times 64$ & \XSolidBrush & 9.0913 & 66.9413 & 0.8191 \\
        \Checkmark & $128\times 128$ & \XSolidBrush & 9.2169 & 56.9813 & 0.8535 \\
        \Checkmark & $256\times 256$ & \XSolidBrush & 9.3046 & 58.9413 & 0.8199 \\
        \Checkmark & $256\times 256$ & \Checkmark & 9.4039 & 57.0631 & 0.8517 \\
        \bottomrule[1pt]
    \end{tabular}}
    \label{tab:style}
\end{table}

Table~\ref{tab:style} shows that removing re-normalization substantially degrades IS, FID, and CLIP-I. The $64\times64$ setting performs best, suggesting that learning with a stronger low-resolution bottleneck is beneficial; higher training resolutions weaken this bottleneck and make the style pathway more prone to reference-structure reliance. The CLIP-encoder variant does not outperform crop tokens, suggesting that a stronger generic visual representation is not necessarily a better style representation here.

\section{Discussion}

\subsection{Methodology Comparisons}
\textbf{Relation to CSGO~\cite{csgo}. }Both CSGO and {\mname} are end-to-end tuning methods for diffusion stylization. The main difference is the supervision and conditioning design. CSGO relies on pre-collected content-style-target triplets, whereas {\mname} constructs pseudo training instances from unpaired single images. Architecturally, CSGO uses an image encoder for style extraction and a ControlNet-style branch for content control. {\mname} instead uses low-resolution crop tokens for style, structural maps for content, and parameter-efficient LoRA/projection layers~\cite{lora}, reducing dependence on auxiliary encoders and triplet construction.

\textbf{Relation to StyleShot~\cite{styleshot}. }StyleShot also aims to disentangle content and style, but does so through dedicated pretrained encoders and a two-stage training strategy. This design is practical for arbitrary style transfer, but the style representation still relies on learned encoder capacity and can retain reference identity, object semantics, or layout cues. {\mname} pursues the same separation objective through data construction and token design rather than encoder capacity: content is represented by explicit structural conditions, while style is represented by low-resolution appearance tokens. This design avoids learning a separate style encoder and addresses content leakage through the scale of the style signal.

\textbf{Limitations.} Because {\mname} deliberately compresses the style reference into low-resolution tokens, it is designed primarily for transferable local appearance statistics rather than exact global composition. It may therefore under-transfer styles whose identity depends on symbolic motifs, legible text, object-level shapes, or precise layout. This is the main trade-off of the proposed bottleneck: reducing reference-structure leakage can also suppress non-local style cues. The method also inherits errors from structural preprocessing, where inaccurate Canny edges or SAM masks can weaken content preservation. These limitations suggest future work on adaptive multi-scale style tokens and more robust structural conditioning.

\section{Conclusion}
We presented {\mname} (\methodfull), a style-encoder-free diffusion stylization framework based on scale-separated conditioning. Instead of extracting style with an auxiliary visual encoder or relying on aligned triplets, {\mname} learns style tokens from stochastic low-resolution crops of unpaired single images. With target-derived structural cues, re-normalization, and skip fusion, {\mname} preserves target structure while transferring reference appearance. Cross-image comparisons, matched controls, and ablations indicate that the crop-token bottleneck reduces reference-layout leakage while retaining transferable style cues; the automatic metrics are used as diagnostics rather than complete definitions of stylization quality.

\bibliographystyle{splncs04}
\bibliography{main}

\end{document}